\documentclass[10pt,twocolumn,letterpaper]{article}

\usepackage{cvpr}
\usepackage{times}
\usepackage{epsfig}
\usepackage{graphicx}
\usepackage{amsmath}
\usepackage{amssymb}

\usepackage{booktabs}
\usepackage{csvsimple}
\usepackage{siunitx}
\usepackage{float}
\usepackage[caption=false, font=footnotesize]{subfig}

\usepackage[pagebackref=true,breaklinks=true,letterpaper=true,colorlinks,bookmarks=false]{hyperref}

\cvprfinalcopy 

\def\cvprPaperID{4} 

\ifcvprfinal\fi
\begin{document}

\title{SoccerNet: A Scalable Dataset for Action Spotting in Soccer Videos}

\author{Silvio Giancola, Mohieddine Amine, Tarek Dghaily, Bernard Ghanem\\
King Abdullah University of Science and Technology (KAUST), Saudi Arabia\\
{\tt\small silvio.giancola@kaust.edu.sa, maa249@mail.aub.edu, tad05@mail.aub.edu, bernard.ghanem@kaust.edu.sa}
}

\maketitle

\begin{abstract}

In this paper, we introduce \emph{SoccerNet}, a benchmark for action spotting in soccer videos. 
The dataset is composed of 500 complete soccer games from six main European leagues, covering three seasons from 2014 to 2017 and a total duration of 764 hours.
A total of 6,637 temporal annotations are automatically parsed from online match reports at a one minute resolution for three main classes of events (Goal, Yellow/Red Card, and Substitution). 
As such, the dataset is easily scalable.
These annotations are manually refined to a one second resolution by anchoring them at a single timestamp following well-defined soccer rules.
With an average of one event every 6.9 minutes, this dataset focuses on the problem of localizing very sparse events within long videos.
We define the task of \emph{spotting} as finding the anchors of soccer events in a video.
Making use of recent developments in the realm of generic action recognition and detection in video, we provide strong baselines for detecting soccer events.
We show that our best model for classifying temporal segments of length one minute reaches a mean Average Precision (mAP) of 67.8\%.
For the spotting task, our baseline reaches an Average-mAP of 49.7\% for tolerances~$\delta$ ranging from 5 to 60 seconds.
Our dataset and models are available at \href{https://silviogiancola.github.io/SoccerNet}{https://silviogiancola.github.io/SoccerNet}.

\end{abstract}

\vspace{-8pt}
\section{Introduction}
\vspace{-2pt}

Sports is a lucrative sector, with large amounts of money being invested on players and teams. 
The global sports market is estimated to generate an annual revenue of \$91 billion~\cite{GlobalSportsMarket}, whereby the European soccer market contributes about \$28.7 billion~\cite{EuropeanFootballMarket}, from which \$15.6 billion alone come from the Big Five European soccer leagues (EPL, La Liga, Ligue 1, Bundesliga and Serie A)~\cite{BigFiveMarket1,BigFiveMarket2}. 
After merchandising, TV broadcast rights are  the second major revenue stream for a soccer club~\cite{BroadcastingRevenue}. 
Even though the main scope of soccer broadcast is entertainment, such videos are also used by professionals to generate statistics, analyze strategies, and scout new players. 
Platforms such as Wyscout~\cite{wyscout}, Reely~\cite{reely}, and Stats SPORTVU~\cite{sportvu} have made sports analytics their core business and already provide various products for  advanced statistics and highlights.


\begin{figure}[t]
    \centering
    \includegraphics[width=0.48\textwidth]{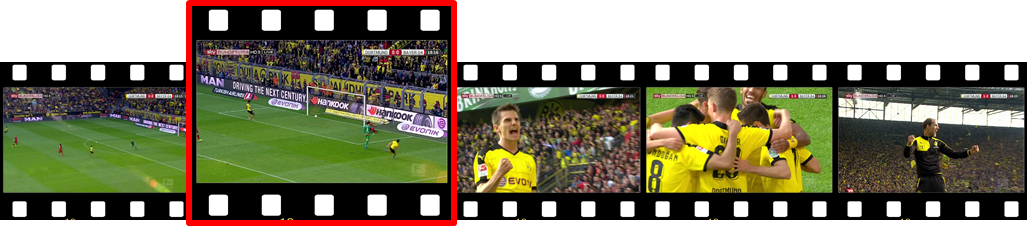}\\ \vspace{1mm} 
    \includegraphics[width=0.48\textwidth]{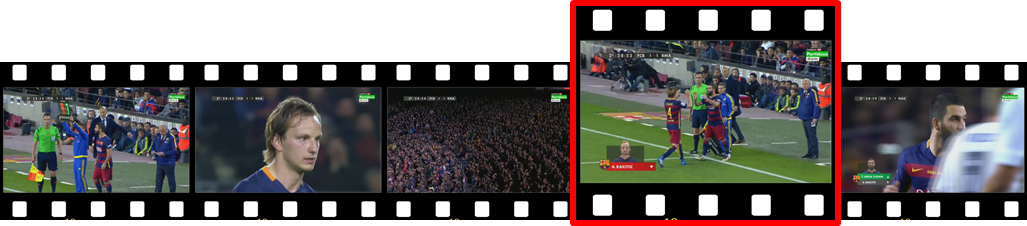}\\ \vspace{1mm}
    \includegraphics[width=0.48\textwidth]{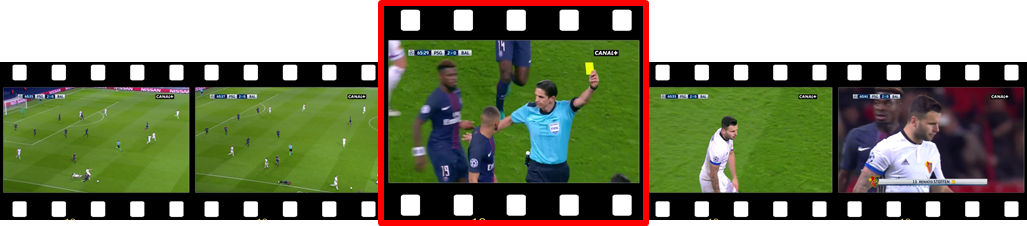}
    \caption{Example of events defined in the context of soccer. 
    From top to bottom: 
    \textbf{Goal}: the instant the ball crosses the goal line.
    \textbf{Substitution}: the instant a players enters the field to substitute an other player.
    \textbf{Card}: the instant the referee shows a card to a player.}
    \vspace{-10pt}
    \label{fig:Teaser}
\end{figure}

In order to get such statistics, professional analysts watch a lot of broadcasts and identify all the events that occur within a game.
According to Matteo Campodonico, CEO of Wyscout, a 400 employee company focusing on soccer data analytics~\cite{wyscout}, it takes over 8 hours to provide up to 2000 annotations per game. 
With more than 30 soccer leagues in Europe, the number of games is very large and requires an army of annotators. 
Even though Amazon Mechanical Turk (AMT) can provide such workforce, building an annotated dataset of soccer games comes at a significant cost.

Automated methods for sports video understanding can help in the localization of the salient actions of a game. 
Several companies such as Reely~\cite{reely} are trying to build automated methods to understand sports broadcasts and would benefit from a large-scale annotated dataset for training and evaluation. 
Many recent methods exist to solve generic human activity localization in video focusing on sports~\cite{bettadapura2016leveraging,caba2017scc,Felsen_2017_ICCV,KarpathyCVPR14}.
However, detecting soccer actions is a difficult task due to the sparsity of the events within a video. 
Soccer broadcast understanding can thus be seen as a sub-problem of video understanding, focusing on a vocabulary of sparse events defined within its own context.

\vspace{4pt}\noindent\textbf{Contributions.}
\textbf{(i)} We propose the task of \emph{event spotting} within a soccer context. 
We define \emph{events} as actions anchored to a single timestamp in a video and, thus, proceed to define and study the task of \emph{spotting} events within soccer videos (Section~\ref{sec:Spotting}).
\textbf{(ii)} We propose \emph{SoccerNet}, a scalable dataset for soccer video understanding. It contains 764 hours of video and 6,637 instances split in three classes (Goal, Yellow/Red Card, and Substitution), which makes it the largest localization dataset in term of total duration and number of instances per class (Section~\ref{sec:DataCollection}). 
\textbf{(iii)} We provide baselines for our dataset in the tasks of video chunk classification and event spotting. Our \emph{minute classifier} reaches a performance of 67.8\% (mAP) and our \emph{event spotter} an Average-mAP of 49.7\% (Section~\ref{sec:Experiments}).

\section{Related Work}

This paper relates to the topics of Sports Analytics, Activity Recognition and Action Localization Datasets.
We give a brief overview of work relevant to each of these topics and highlight how our paper contributes to each of them.

\vspace{4pt}\noindent\textbf{Sports Analytics.~~} Many automated sports analytics methods have been developed in the computer vision community to understand sports broadcasts~\cite{bettadapura2016leveraging,d2010review,Felsen_2017_ICCV,kapela2014real,ramanathan2016detecting}. 
They produce statistics of events within a game by either analyzing camera shots or semantic information. 
Ekin \etal~\cite{ekin2003automatic} present a cornerstone work for game summarization based on camera shot segmentation and classification, followed by Ren \etal~\cite{ren2005football} who focus on identifying video production patterns. 
Huang \etal~\cite{huang2006semantic} analyze semantic information to automatically detect goals, penalties, corner kicks, and card events. 
Tavassolipour \etal~\cite{tavassolipour2014event} use Bayesian networks to summarize games by means of semantic analysis.

More recent work in this category focuses on deep learning pipelines to localize salient actions in soccer videos. 
Baccouche \etal~\cite{baccouche2010action} use a Bag-of-Words (BOW) approach with SIFT features to extract visual content within a frame. 
They use such representations to train a Long Short Term Memory (LSTM) network that temporally traverses the video to detect the main actions. 
Jiang \etal~\cite{jiang2016automatic} propose a similar methodology using Convolution Neural Networks (CNN) to extract global video features rather than local descriptors. 
They also use a play-break structure to generate candidate actions. 
Tsagkatakis \etal~\cite{tsagkatakis2017goal} present a two-stream approach to detect goals, 
while Homayounfar \etal~\cite{Homayounfar_2017_CVPR} recently present a deep method for sports field localization, which is crucial for video registration purposes.

The main impediment for all these works is the lack of reference datasets/benchmarks that can be used to evaluate their performance at large-scale and standardize their comparison. They all use small and custom-made datasets, which contain a few dozen soccer games at most. We argue that intelligent sports analytics solutions need to be scalable to the size of the problem at hand. Therefore, to serve and support the development of such scalable solutions, we propose a very large soccer-centric dataset that can be easily expanded and enriched with various types of annotations.

\vspace{4pt}\noindent\textbf{Activity Recognition.~~} 
Activity recognition focuses on understanding videos by either detecting activities or classifying segments of video according to a predefined set of human-centric action classes. 
A common pipeline consists of proposing temporal segments~\cite{Buch_2017_CVPR,caba2016fast,Gao_2017_ICCV,shou2016temporal}, which are in turn further pruned and classified~\cite{girdhar2017actionvlad,wang2015action}.
Common methods for activity classification and detection make use of 
dense trajectories~\cite{van2015apt,wang2013action,wang2015action,wang2016temporal}, 
actionness estimation~\cite{chen2014actionness,Gao_2017_ICCV,Zhao_2017_ICCV}, 
Recurrent Neural Networks (RNN)~\cite{sstad_buch_bmvc17,Buch_2017_CVPR,escorcia2016daps},
tubelets~\cite{Kalogeiton_2017_ICCV, Saha_2017_ICCV},
and handcrafted features~\cite{caba2016fast,mettes2015bag,yu2015fast}.

In order to recognize or detect activities within a video, a common practice consists of \textbf{aggregating} local features and \textbf{pooling} them, looking for a consensus of characteristics~\cite{KarpathyCVPR14,simonyan2014two}. 
While naive approaches use mean or maximum pooling, 
more elaborate techniques such as 
Bag-of-Words (BOW)~\cite{csurka2004visual,sivic2003video}, 
Fisher Vector (FV)~\cite{jaakkola1999exploiting,perronnin2007fisher,perronnin2010improving}, and
VLAD~\cite{arandjelovic2013all,jegou2010aggregating} 
look for a structure in a set of features by clustering and learning to pool them in a manner that improves discrimination.
Recent works extend those pooling techniques by incorporating them into Deep Neural Network (DNN) architectures, namely 
NetFV~\cite{lev2016rnn,perronnin2015fisher,sydorov2014deep}, 
SoftDBOW~\cite{philbin2008lost}, and 
NetVLAD~\cite{arandjelovic2016netvlad,girdhar2017actionvlad}.
By looking for correlations between a set of primitive action representations, ActionVLAD~\cite{girdhar2017actionvlad} has shown state-of-the-art performance in several activity recognition benchmarks.

To further improve activity recognition, recent works focused on exploiting \textbf{context}~\cite{caba2017scc,Dai_2017_ICCV,miech2017learnable}, which represent and harness information in both temporal and/or spatial neighborhood, or on \textbf{attention}~\cite{nguyen2015stap}, which learns an adaptive confidence score to leverage this surrounding information. 
In this realm, Caba Heilbron \etal~\cite{caba2017scc} develop a semantic context encoder that exploits evidence of objects and scenes within video segments to improve activity detection effectiveness and efficiency. Miech \etal~\cite{miech2017learnable}, winners of the first annual Youtube 8M challenge~\cite{abu2016youtube}, show how learnable pooling can produce state-of-the-art recognition performance on a very large benchmark, when recognition is coupled with context gating. More recently, several works use temporal context to localize activities in videos~\cite{Dai_2017_ICCV} or to generate proposals~\cite{Gao_2017_ICCV}. Furthermore, Nguyen \etal~\cite{nguyen2015stap} present a pooling method that uses spatio-temporal attention for enhanced action recognition, while Pei \etal~\cite{Pei_2017_CVPR} use temporal attention to gate neighboring observations in a RNN framework. Note that attention is also widely used in video captioning~\cite{Hori_2017_ICCV,Krishna_2017_ICCV,Mazaheri_2017_ICCV}.

Activity recognition and detection methods are able to provide good results for these complicated tasks. However, those methods are based on DNNs and require large-scale and rich datasets to learn a model. By proposing a large-scale dataset focusing on event spotting and soccer, we encourage algorithmic development in those directions. 

\vspace{4pt}\noindent\textbf{Datasets.~~} 
Multiple datasets are available for video understanding, especially for video classification. They include 
\textbf{Hollywood2}~\cite{marszalek2009actions} and
\textbf{HMDB}~\cite{Kuehne11}, both focusing on movies;
\textbf{MPII Cooking}~\cite{rohrbach2012database}, focusing on cooking activities;
\textbf{UCF101}~\cite{soomro2012ucf101}, for classification in the wild;
\textbf{UCF Sports}~\cite{rodriguez2008action},
\textbf{Olympics Sports}~\cite{niebles2010modeling} and
\textbf{Sports-1M}~\cite{KarpathyCVPR14}, focusing on sports;
\textbf{Youtube-8M}~\cite{abu2016youtube} and
\textbf{Kinetics}~\cite{kay2017kinetics}, both tackling large scale video classification in the wild. 
They are widely used in the community but serve the objective of video classification rather than activity localization.

The number of benchmark datasets focusing on action localization is much smaller.
\textbf{THUMOS14}~\cite{THUMOS14} is the first reasonably scaled benchmark for the localization task with a dataset of 413 untrimmed videos, totaling 24 hours and 6,363 activities, split into 20 classes. 
\textbf{MultiTHUMOS}~\cite{yeung2015every} is a subset of THUMOS, densely annotated for 65 classes over unconstrained internet videos. 
\textbf{ActivityNet}~\cite{caba2015activitynet} tackles the issue of general video understanding using a semantic ontology, proposing challenges in trimmed and untrimmed video classification, activity localization, activity proposals and video captioning. ActivityNet 1.3 provides a dataset of 648 hours of untrimmed videos with 30,791 activity candidates split among 200 classes. It is so far the largest localization benchmark in terms of total duration. 
\textbf{Charades}~\cite{sigurdsson2016hollywood} is a recently compiled benchmark for temporal activity segmentation that crowd-sources the video capturing process. After collecting a core set of videos from YouTube, they use AMT to augment their data by recording them at home. This dataset consists of a total of 9,848 videos and 66,500 activities.
More recently, Google proposed \textbf{AVA}~\cite{gu2017ava} as a dataset to tackle dense activity understanding. They provide 57,600 clips of 3 seconds duration taken from featured films, annotated with 210,000 dense spatio-temporal labels across 100 classes, for a total of 48 hours of video. While the main task of AVA is to classify these 3 seconds segments, such dense annotation can also be used for detection.

Within the multimedia community, \textbf{TRECVID} has been the reference benchmark for over a decade~\cite{awad2016trecvid,smeaton2006evaluation}.
They host a ``Multimedia Event Detection'' (MED) and a ``Surveillance Event Detection'' (SED) task every year, using the \textbf{HAVIC} dataset~\cite{strassel2012creating}.
These tasks focus on finding all clips in a video collection that contain a given event, with a textual definition, in multimedia and surveillance settings. Also, Ye \etal~\cite{ye2015eventnet} propose \textbf{EventNet}, a dataset for event retrieval based on a hierarchical ontology, similar to ActivityNet.
We argue that these two datasets both focus on large-scale information retrieval rather than video understanding.

\begin{figure}[t]
    \centering
    \includegraphics[width=0.48\textwidth]{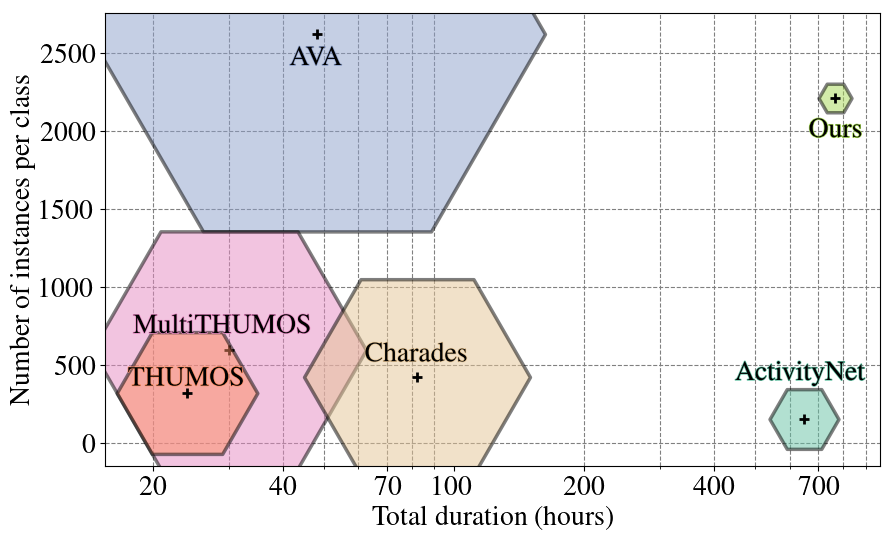}
    \caption{
    Dataset comparison in term of number of instance per class, and total duration.
    The size of the hexagon shows the density of the event within the video.
    Our dataset has the largest amount of instances per class and the largest total duration, despite being sparse, which makes the task of localization more difficult.}
    \label{fig:DatasetsComparison}
\end{figure}

We propose \emph{SoccerNet}, a scalable and soccer-focused dataset for event spotting. It contains 500 games, 764 hours of video and 6,637 instances split in three classes (Goal, Yellow/Red Card, and Substitution), which makes it one of the largest dataset in term of total duration and number of instances per class. With an average of one event every 6.9 minutes, our dataset has a sparse distribution of events in long untrimmed videos, which makes the task of localization more difficult. The annotations are obtained within one minute at no cost by parsing sports websites, and further refined in house to one second resolution. We define our dataset as easily scalable since annotations are obtained for \emph{free} from online match reports.
Table~\ref{tab:DatasetsComparison} shows a breakdown description and comparison of the datasets available for the problem of action localization. Figure~\ref{fig:DatasetsComparison} shows a graphical comparison between these datasets in terms of the number of instances per class and the total duration of videos they contain.

\begin{table*}[htb]
    \centering
    \caption{Comparison of benchmark datasets currently tackling the task of action localization.}
    \vspace{5pt}
    \begin{filecontents*}{img/Dataset/datasets.csv}
Dataset,Context,NumVideo,Duration,NumActivity,NumClass,SparsityVideo,SparsityHour,SparsityMin,AnnotSource,ActivityClass
THUMOS'14~\cite{THUMOS14}               ,General,413,24,6363,20,15.407,260.4,0.23,InHouse,318
MultiTHUMOS~\cite{yeung2015every},General,400,30,38690,65,96.725,1289.7,0.047,OutSourcing,595
ActivityNet~\cite{caba2015activitynet} ,General,19994,648,30791,200,1.54,47.5,1.263,AMT/OutSourcing,154
Charades~\cite{sigurdsson2016hollywood},General,9848,82,66500,157,6.753,811,0.074,AMT,424
AVA~\cite{gu2017ava},Movies,57600,48,210000,80,3.646,4375,0.014,InHouse,2625
Ours                                    ,Soccer,1000,764,6637,3,6.637,8.7,6.907,Webly+InHouse,2212
\end{filecontents*}
\csvreader[tabular=l||c|r|r|r|r|r|r, 
	table head=  \textbf{Dataset} &  \textbf{Context}  &  \textbf{\#Video} &  \textbf{\#Instance} &  \textbf{Duration} &  \multicolumn{1}{c|}{\textbf{Sparsity}} &  \textbf{Classes} &  \multicolumn{1}{c}{\textbf{Instance}} \\
	\textbf{} &  \textbf{}  &  \textbf{} &  \textbf{} &  \multicolumn{1}{c|}{\textbf{(hrs)}} &   \textbf{(event/hr)}  &  \textbf{} &  \textbf{per class}  \\\midrule,
	late after line=\ifthenelse{\equal{\Dataset}{Ours}}{\\\midrule}{\\}]
	{img/Dataset/datasets.csv}%
	{Dataset=\Dataset, Context=\Context, NumVideo=\NumVideo, NumActivity=\NumActivity, Duration=\Duration, SparsityVideo=\SparsityVideo, SparsityHour=\SparsityHour, AnnotSource=\AnnotSource, NumClass=\NumClass, ActivityClass=\ActivityClass}%
	{\textbf{\Dataset} & 
	\Context & 
	\NumVideo & 
	\NumActivity &
	\Duration & 
	\SparsityHour & 
	\NumClass & 
	\ActivityClass }
    \label{tab:DatasetsComparison}
\end{table*}

\section{Spotting Sparse Events in Soccer}
\label{sec:Spotting}

In this context, we define the concept of \emph{events} and the task of \emph{spotting events} within soccer videos.

\vspace{4pt}\noindent\textbf{Events:~~}
Sigurdsson \etal~\cite{Sigurdsson_2017_ICCV} recently question the concept of temporal boundaries in activities. They re-annotated Charades~\cite{sigurdsson2016hollywood} and MultiTHUMOS~\cite{yeung2015every} (using AMT), and conclude that the average agreement with the ground truth is respectively 72.5\% and 58.8\% tIoU. This clearly indicates that temporal boundaries are ambiguous. However, Sigurdsson \etal~\cite{Sigurdsson_2017_ICCV} observe that central frames within an activity offer more consensus among annotators.

Chen \etal~\cite{chen2014actionness} define the concept of \emph{action} and \emph{actionness} by underlining 4 necessary aspects that define an action: an agent, an intention, a bodily movement, and a side-effect. Dai \etal~\cite{Dai_2017_ICCV} define an \emph{activity} as a set of events or actions, with a beginning and an ending time. In our work, we define the concept of \emph{event} as an action that is anchored in a single time instance, defined within a specific context respecting a specific set of rules.  We argue that defining every action with temporal boundaries is ambiguous for multiple reasons:

\vspace{-5pt}



\begin{enumerate}
    \item An action can occur in a glimpse, such as \emph{``a man dropped his wallet''} or \emph{``a man put a letter in the mail''}. While there are no well-defined boundaries for such actions, a sole instant can readily define these events.
\vspace{-5pt}

    \item An action can be continuous within a live video, hence it is unclear when it starts or stops. For instance, time boundaries in video for actions such as \emph{``the night is falling''} or \emph{``the ice is melting in my glass''}, rely on a subjective discrimination between measurable quantities such as the illumination level or visual changes in matter state. 
\vspace{-5pt}

    \item An action can overlap and conflict with another. Consider a video of a man walking his dog, when he suddenly receives a phone call. 
    It is not clear whether the activity \emph{``taking a phone call''} actually cancels out the activity \emph{``walking a dog''}, or the activity \emph{``walking a dog''} should be split into two parts as opposed to one single segment overlapping the \emph{``taking a phone call''} instance.
    \vspace{-5pt}

\end{enumerate}

Current benchmarks such as THUMOS14~\cite{THUMOS14}, ActivityNet~\cite{caba2015activitynet}, and Charades~\cite{sigurdsson2016hollywood} only focus on activities with temporal boundaries and cope with ambiguities by anchoring an activity with a consensus between several annotators. 
This ambiguity motivates the recently developed AVA~\cite{gu2017ava} dataset that attempts to tackle the atomic characteristic of actions by providing dense fine-scale annotations within a short time duration (3 seconds).

In the multimedia community, the concept of events is generally more vague and overlaps with the concept of actions and activities.
In the MED task of the TRECVID benchmark~\cite{awad2016trecvid}, an event is defined as a kit which consists of a \emph{mnemonic title} for the event, a \emph{textual definition}, an \emph{evidential description} that indicates a non-exhaustive list of textual attributes, and a set of \emph{illustrative video examples}.
They propose a specific set of events, providing a description and defining rules for the start and end times. Such work underlines our hypothesis that events need to be defined with a set of rules and within specific circumstances.

In the context of live soccer broadcasts, it is unclear when a given action such as \emph{``scoring a goal''} or \emph{``making a foul''} starts and stops.
For similar reasons, the beginning and end of activities such as \emph{``scoring a 3 points shot''} or a \emph{``slam dunk''} in a basketball broadcast are ambiguous.
We argue that sports respect well-established rules and define an action vocabulary anchored in a single time instance. 
In fact, soccer rules provide a strict definition of \emph{``goal''}, \emph{``foul''}, \emph{``card''}, \emph{``penalty kick''}, \emph{``corner kick''}, \etc and also anchor them within a single time.
Similarly, Ramanathan \etal~\cite{ramanathan2016detecting} define the action \emph{``basket-ball shot''} as a 3 seconds activity and its ending time as the moment the ball crosses the basket.
Defining starting or stopping anchors around such events or fixing its duration would be considered as subjective and biased by the application.

\vspace{4pt}\noindent\textbf{Spotting:~~}
Rather than identifying the boundaries of an action within a video and looking for similarities within a given temporal Intersection-over-Union (tIoU), we introduce the task of \emph{spotting}.
Spotting consists of finding the anchor time (or \emph{spot}) that identifies an \emph{event}. Intuitively, the closer the candidate spot is from the target, the better the spotting is, and its capacity is measured by its distance from the target. Since perfectly spotting a target is intrinsically arduous, we introduce a tolerance $\delta$ within which a event is considered to be spotted (\emph{hit}) by a candidate.
We believe that event spotting is better defined and easier than detection since it focuses only on identifying the presence of an event within a given tolerance. An iterative process can refine such tolerance at will by using fine localization methods around candidate spots. 

By introducing the task of spotting, we also define the metric to be used for evaluation.
First of all, we define a candidate spot as positive if it lands within a tolerance $\delta$ around the anchor of an event. For each tolerance, we can recast the spotting problem as a general temporal detection problem, where the tIoU threshold used is very small. In that case, we can compute the recall, precision, Average Precision (AP) for each given class, and a mean Average Precision (mAP) across all classes. For general comparison, we also define an Average-mAP over a set of predefined $\delta$ tolerances, in our case below the minute. 




\section{Data Collection}
\label{sec:DataCollection}
We build our dataset in three steps: 
\textbf{(i)} we collect videos from online sources; 
\textbf{(ii)} we synchronize the game and video times by detecting and reading the game clock; 
and \textbf{(iii)} we parse match reports available in the web to generate temporal aligned annotations.

\subsection{Collecting Videos}
We compile a set of 500 games from the main European Championships during the last 3 seasons as detailed in Table~\ref{tab:DatasetVideoCollected}. Each game is composed of 2 untrimmed videos, one for each half period. The videos come from online providers, in a variety of encodings (MPEG, H264), containers (MKV, MP4, and TS), frame rates (25 to 50 fps), and resolutions (SD to FullHD). The dataset consumes almost 4TB, for a total duration of 764 hours. The games are randomly split into 300, 100, and 100 games for training, validation, and testing ensuring similar distributions of the events between the classes and the datasets.

\begin{table}[htb]
	\centering
	\caption{Summary of the video collection for our dataset.}
	\vspace{5pt}
	\begin{filecontents*}{img/Dataset/games_crop224.csv}
Championship,seasA,seasB,seasC,Total
EN - EPL,6,49,40,95
ES - LaLiga,18,36,63,117
FR - Ligue 1,1,3,34,38
DE - BundesLiga,8,18,27,53
IT - Serie A,11,9,76,96
EU - Champions,37,45,19,101
Total,81,160,259,500
\end{filecontents*}
\csvreader[tabular=l||c|c|c||>{\bfseries}c, 
	table head=   &  \multicolumn{3}{c||}{Seasons} &   \\
	League &  14/15 &  15/16 &  16/17 &  Total \\\midrule,
	late after line=\ifthenelse{\equal{\Championship}{Total}}{\\\midrule}{\\}]
	{img/Dataset/games_crop224.csv}%
	{Championship=\Championship,
	seasA=\seasA,
	seasB=\seasB,
	seasC=\seasC,
	Total=\Total}%
	{\ifthenelse{\equal{\Championship}{Total}}
		{\textbf{\Championship}}{\Championship} & 
		\ifthenelse{\equal{\Championship}{Total}}
		{\textbf{\seasA}}{\seasA} & 
		\ifthenelse{\equal{\Championship}{Total}}
		{\textbf{\seasB}}{\seasB} & 
		\ifthenelse{\equal{\Championship}{Total}}
		{\textbf{\seasC}}{\seasC} & 
		\ifthenelse{\equal{\Championship}{Total}}
		{\textbf{\Total}}{\Total} 
	}
	\label{tab:DatasetVideoCollected}
\end{table}

\vspace{-15pt}

\subsection{Game Synchronization with OCR}\label{subsec:OCR}
The video of the games are untrimmed and contains spurious broadcast content before and after the playing time. Finding a mapping between the game time and the video time is necessary to align the annotations from the web sources to the videos. Soccer games have a continuous game flow, \ie the clock never stops before the end of a half, hence there is a simple linear relationship (with slope 1) between the video and the game time. Wang \etal~\cite{wang2017soccer} propose a method using the center circle of the field and the sound of the referee whistle to identify the start of the game. We argue that focusing the effort on a single instant is prone to error. In contrast, we focus on detecting the game clock region within multiple video frames and identify the game time through Optical Character Recognition (OCR) at different instants.

The clock is displayed in most of the frames throughout the video, though its shape and position vary between leagues. We leverage a statistical study of the pixel intensity deviation within a set of $N$ random frames to identify candidates for the clock region. We run the Tesseract OCR Engine~\cite{smith2007overview} on the candidate clocks and look for a coherent time format for each of the $N$ frames. To cope with eventual misreadings in the clock, we use a RANSAC~\cite{fischler1981random} approach to estimate the linear relation between the game clock and the video time, enforcing a unitary gradient to our linear model. Our method also checks for the temporal integrity of the video, reporting temporal inconsistencies. To verify the quality of this game-to-video temporal alignment, we manually annotate the start of the game for all 500 games and report an accuracy of 90\% for automatically estimating the start of both halves within a tolerance of two seconds, using a set of $N=200$ frames.

\subsection{Collecting Event Annotations}
For our dataset, we obtain event annotations for \emph{free} by parsing match reports provided by league websites\footnote{We choose \url{www.flashscore.info} to get our annotations since they provide a wide number of summaries and have a consistent format across their match reports.}. They summarize the main actions of the game and provide the minute at which the actions occur. We categorize these events into our three main categories: \emph{``goals''}, \emph{``cards''} and \emph{``substitutions''}. We  parse and mine the annotations for all games of the Big Five European leagues (EPL, La Liga, Ligue  1, Bundesliga and Serie A) as well as the Champions League from 2010 to 2017, for a total of 171,778 annotations corresponding to 13,489 games. 
For sake of storage, we focus on our subset of videos for the 500 games and use only 6,637 events. To resolve these free annotations to one second level, we manually annotate each event within one second resolution by first retrieving its minute annotation and refining it within that minute window. To do so, we define the temporal anchors for our events from their definitions within the rules of soccer. We define a \emph{``goal''} event as the instant the ball crosses the goal line to end up in the net. We define the \emph{``card''} event as the instant the referee shows a player a yellow or a red card because of a foul or a misbehaviour. Finally, we define the \emph{``substitution''} event as the instant a new player enters in the field. 
We ensure those definition were coherent when annotating the dataset.
Apart for the substitutions that occur during half time break, almost all of our instances follow their definitions.

\subsection{Dataset Scalability}
We believe that scaling our dataset is cheap and easy, since web annotations are freely available with one minute resolution.
Algorithm can either use the weakly annotated events within one minute resolution or generate a complete one second resolution annotation which is estimated to take less than 10 minutes per game.
We also argue that broadcast providers can easily scale up such datasets by simply providing more videos and richer annotations. 

\section{Experiments}
\label{sec:Experiments}

We focus the attention of our experiments on two tasks: event classification for chunks of one minute duration, and event spotting within an entire video. 
For these tasks, we report and compare the performance metrics for various baseline methods when trained on weakly annotated data (\ie one minute resolution) and the improvement that is achieved by training on one second resolution annotations.

\subsection{Video Representation}

Before running any experiments, we extract C3D~\cite{tran2015learning}, I3D~\cite{carreira2017quo}, and ResNet~\cite{he2016deep} features from our videos to be used by our baselines.
The videos are trimmed at the game start, resized and cropped at a $224\times 224$ resolution, and unified at 25fps. 
Such representation guarantees storage efficiency, fast frame access, and compatible resolution for feature extraction. 
\textbf{C3D}~\cite{tran2015learning} is a 3D CNN feature extractor that stacks 16 consecutive frames and outputs at the \emph{fc7} layer a feature of dimension 4,096. It is pretrained on Sport-1M~\cite{KarpathyCVPR14}. 
\textbf{I3D}~\cite{carreira2017quo} is based on Inception~V1~\cite{szegedy2016rethinking}, uses 64 consecutive frames, and is pretrained on Kinetics~\cite{kay2017kinetics}. In this work, we only extract the RGB features at the \emph{PreLogits} layer of length 1,024 so to maintain a reasonable computational runtime.
They have been shown to produce only meager improvements when flow features are used~\cite{carreira2017quo}.
\textbf{ResNet}~\cite{he2016deep} is a very deep network that outputs a 2,048 feature representation  per frame at the \emph{fc1000} layer. In particular, we use ResNet-152 pretrained on ImageNet~\cite{deng2009imagenet}. Since ResNet-152 applies to single images, it does not intrinsically embed contextual information along the time axis. 
We use TensorFlow~\cite{tensorflow2015-whitepaper} implementations to extract features every 0.5 second (s).
In order to simplify and unify the dimension of those features, we reduce their dimensionality by applying Principal Component Analysis (PCA) on the 5.5M features we extract per model.
We reduce C3D, I3D, and ResNet-152 features to a dimension of 512 and respectively maintain 94.3\%, 98.2\%, and 93.9\% of their variance. 
For the benchmark purpose, we provide the original and cropped versions of the videos, as well as the original and the reduced versions of all the features extracted every 0.5s.

\subsection{Video Chunk Classification}
\label{subsec:Classifier}

Similar to the setup in the AVA dataset~\cite{gu2017ava}, localization can be cast as a classification problem for densely annotated chunks of video, especially since we gather webly annotations. We split our videos into chunks of duration 1 minute, annotated with all events occurring within this minute, gathering respectively 1246, 1558, 960 and 23750 chunks for cards, substitutions, goals and backgrounds for the training dataset, 115 of which having multiple labels. We aggregate the 120 features within a minute as input for different versions of shallow pooling neural networks. By using a sigmoid activation function at the last layer of these networks, we allow for multi-labelling across the candidates. We use an Adam optimizer that minimizes a multi binary cross-entropy loss for all the classes.
We used a step decay for the learning rate and an early stopping technique based on the validation set performances.
Following best practices in the field, the evaluation metric in this case is mAP (classification) across the three classes on the designated testing set.
In what follows, we report strong baseline results using different video features, different pooling techniques, and compare solutions to cope with the imbalanced dataset.

\vspace{4pt}\noindent\textbf{Learning How to Pool:~~}
We investigate the usage of different feature representations and various pooling methods. We propose shallow neural networks that handles the input matrix of dimension $120\times512$. We test a \textbf{mean} and a \textbf{max pooling} operation along the aggregation axis that output 512-long features. We use a custom \textbf{CNN} with a kernel of dimension $512\times20$ that traverses the temporal dimension to gather temporal context. Finally, we use implementations of \textbf{SoftDBOW}, \textbf{NetFV}, \textbf{NetVLAD} and \textbf{NetRVLAD} provided by Miech \etal~\cite{miech2017learnable}, who leverage a further context-gating layer. After each of these pooling layer, we stack a fully connected layer with a dropout layer (keep probability 60\%) that predicts the labels for the minutes of video and prevent overfitting.

Table~\ref{tab:PoolingFeatures} summarizes a performance comparison between the various pooling methods when applied to the testing set. 
First of all, we notice similar results across features by using mean and max pooling, that only rely on a single representation of the set of 120 features and not its distribution.
Using the custom CNN layer, which is an attempt to gather temporal context, ResNet-152 performs better than C3D which performs better than I3D. We believe that the I3D and C3D already gather temporal information for 64 and 16 frames.

We can notice that the gap between the features increases by using the pooling methods proposed by Miech \etal~\cite{miech2017learnable}, which is a way to embed context along the temporal dimension. We believe that I3D and C3D features already rely on a temporal characterization within the stack of frames. On the other hand, the ResNet-152 provides a representation that focuses only on the spatial aspect within a frame. We believe that the temporal pooling methods provides more redundant information for I3D and C3D, than for ResNet-152. For this reason, we argue that ResNet-152 features provide better results when coupled with any temporal pooling methods provided by Miech \etal~\cite{miech2017learnable}.

Focusing on the pooling, VLAD-based methods are at the top of the ranking, followed by the deep versions of the FV and BoW methods. Such improvement is attributed to the efficient clustering for the 120 features learned in NetVLAD~\cite{arandjelovic2016netvlad} providing state-of-the-art results for action classification~\cite{girdhar2017actionvlad}. Note that NetRVLAD performs similarly if not better than NetVLAD by relying only on the average and not the residuals for each clustering, reducing the computational load~\cite{miech2017learnable}. For the rest of the experiment we are relying exclusively on ResNet-152 features.


\begin{table}[htb]
	\centering
	\caption{Classification metric (mAP) for different combinations of frame representations and pooling methods.}
	\vspace{5pt}
	\begin{filecontents*}{img/Results/Results_Network_Features.csv}
Pooling,IED,CED,ResNet
Mean Pool.,40.8,40.7,40.2
Max Pool.,50.1,52.4,52.4
CNN,44.1,47.8,53.5
SoftDBOW,46.3,56.9,58.9
NetFV,44.7,59.6,64.4
NetRVLAD,41.3,59.9,\textbf{65.9}
NetVLAD,43.4,60.3,65.2
\end{filecontents*}
\csvreader[tabular=l||c|c|c, 
	table head=   &  \multicolumn{3}{c}{\textbf{Frame features}}  \\
	\textbf{Pooling} &  \textbf{I3D} &  \textbf{C3D} &  \textbf{ResNet}  \\\midrule,
	late after line=\\]
	{img/Results/Results_Network_Features.csv}%
	{Pooling=\Pooling,
	IED=\IED,
	CED=\CED,
	ResNet=\ResNet}%
	{\textbf{\Pooling} & 
	~~\IED~~ & 
	~~\CED~ & 
	~~\ResNet~~}
	\label{tab:PoolingFeatures}
	\vspace{-4mm}
\end{table}

For the various pooling methods, the number of clusters can be fine-tuned. In Table~\ref{tab:PoolingFeatures}, we use $k=64$ clusters, which can be interpreted as the vocabulary of atomic elements that are learned to describe the events.
Intuitively, one can expect that a richer and larger vocabulary can enable better overall performance~\cite{girdhar2017actionvlad}. We show in Table~\ref{tab:PoolingClusters} that this intuition is true within a certain range of values $k$, beyond which the improvement is negligible and overfitting occurs. The performance of all pooling methods seem to plateau when  more than 256 clusters are used for the quantization. The best results are registered when NetVLAD is used with 512 clusters. Nevertheless, the computational complexity increases linearly with the number of clusters, hence computational times grow drastically. 


\begin{table}[htb]
	\centering
	\caption{Classification metric (mAP) for different number of cluster for the pooling methods proposed by Miech \etal~\cite{miech2017learnable}.}
	\vspace{5pt}
	\begin{filecontents*}{img/Results/Results_Network_Cluster2.csv}
K,SOFTBOW,NETFV,RVLAD,VLAD
16,54.9,63.0,64.4,65.2
32,57.7,64.0,63.8,65.1
64,58.8,64.1,65.3,65.2
128,60.6,64.4,67.0,65.6
256,61.3,63.8,67.7,67.0
512,62.0,62.1,67.4,\textbf{67.8}
\end{filecontents*}
\csvreader[tabular=l||c|c|c|c, 
	table head=    &  \multicolumn{4}{c}{\textbf{Pooling Methods}}  \\
	 \textbf{$k$}  &  \textbf{SoftBOW} &  \textbf{NetFV} &  \textbf{NetRVLAD} &  \textbf{NetVLAD}  \\\midrule,
	late after line=\\]
	{img/Results/Results_Network_Cluster2.csv}%
	{K=\K,
	SOFTBOW=\SOFTBOW,
	NETFV=\NETFV,
	RVLAD=\RVLAD,
	VLAD=\VLAD}%
	{\textbf{\K} &
	\SOFTBOW & 
	\NETFV & 
	\RVLAD & 
	\VLAD}
	\label{tab:PoolingClusters}
	\vspace{-4mm}
\end{table}

\vspace{4pt}\noindent\textbf{Coping with Imbalanced Data:~~}
The performance of classifiers are significantly affected when training sets are imbalanced. 
Due to the sparsity of our events, we have numerous background instances. Here, we present three main techniques to cope with this imbalance. 
One method focuses on \textbf{weighting (Weig)} the binary cross-entropy with the ratio of negative samples to enforce the learning of the positive examples. 
Another method applies a \textbf{random downsampling (Rand)} on the highest frequency classes, or by \textbf{hard negative mining (HNM)}, \ie by sampling the examples that are misclassified the most in the previous epoch. 
The third method uses \textbf{Data Augmentation (Augm)} to balance the classes. 
In that case, we use the fine annotation of the event and slide the minute window with a stride of 1s within $\pm$20s of the event spot to sample more video segments for the sparsest event classes.
We argue that a chunk of 1 minute within $\pm$10s around the anchor of the event still contains this event, and the pooling method should be able to identify it.
Although, note that our data augmentation requires the data to be finely annotated.

Table~\ref{tab:Imbalanced} shows the classification mAP for the testing dataset, training with the previous pooling methods on ResNet features, and using the aforementioned strategies to cope dataset imbalance.
We see that weighting slightly improves the metric. 
Both downsampling methods actually lead to the worst results, because of the reduced amount of data the model has been trained on at each epoch.
Using the second resolution annotations to augment the data helps to achieve slightly better classification results.

\begin{table}[htb]
	\centering
	\caption{ Classification metric (mAP) using different solutions to cope with an imbalanced dataset on our pooling methods, using ResNet-152 features.}
	\vspace{5pt}
	\begin{filecontents*}{img/Results/Results_Imbalanced_Pooling.csv}
Imbalance,Nothing,Weighting,Rand,HNM,Augm
Mean Pool.,41.5,42.4,40.9,42.4,43.9
Max Pool.,39.3,52.4,48.3,52.4,51.5
CNN,52.9,52.1,50.5,49.8,53.5
SoftDBOW,58.4,59.7,48.7,48.8,50.8
NetFV,64.2,63.8,58.0,60.5,65.2
NetRVLAD,65.2,64.9,61.4,61.0,\textbf{66.7}
NetVLAD,65.0,64.9,59.4,59.5,65.1
\end{filecontents*}
\csvreader[tabular=l||c|c|c|c|c, 
	table head=  & \multicolumn{5}{c}{\textbf{Imbalance}}  \\
	\textbf{Pooling} & \textbf{Orig} &  \textbf{Weig} &  \textbf{Rand} &  \textbf{HNM} &  \textbf{Augm}  \\\midrule,
	late after line=\\]
	{img/Results/Results_Imbalanced_Pooling.csv}%
	{Imbalance=\Imbalance,
	Nothing=\Nothing,
	Weighting=\Weighting,
	Rand=\Rand,
	HNM=\HNM,
	Augm=\Augm}%
	{\textbf{\Imbalance} & 
	\Nothing & 
	\Weighting & 
	\Rand &
	\HNM & 
	\Augm}
	\label{tab:Imbalanced}
	\vspace{-5mm}
\end{table}

\subsection{Spotting}
In this section, we discuss the task of event spotting in soccer videos. 
We use the models trained in the classifier task and apply them in a sliding window fashion on each testing video, with a stride of 1s, thus, leading to a second resolution score along for each event class.
We investigate the spotting results of three strong baselines
\textbf{(i)} a watershed method to compute segment proposals and use the \textbf{center} time within the segment to define our candidate; 
\textbf{(ii)} the time index of the \textbf{maximum} value of the watershed segment as our candidate; and
\textbf{(iii)} the local maxima along all the video and apply \emph{non-maximum-suppression} (\textbf{NMS}) within a minute window.
The evaluation metric is the mAP with tolerance $\delta$ as defined for spotting in Section~\ref{sec:Spotting}, as well as, the Average-mAP expressed as an area under the mAP curve with tolerance ranging from 5 to 60 seconds.

\begin{figure*}[htb]
    \centering
    \subfloat[\textbf{Model trained on chunks of 60s}]
    {\includegraphics[width=0.32\linewidth]{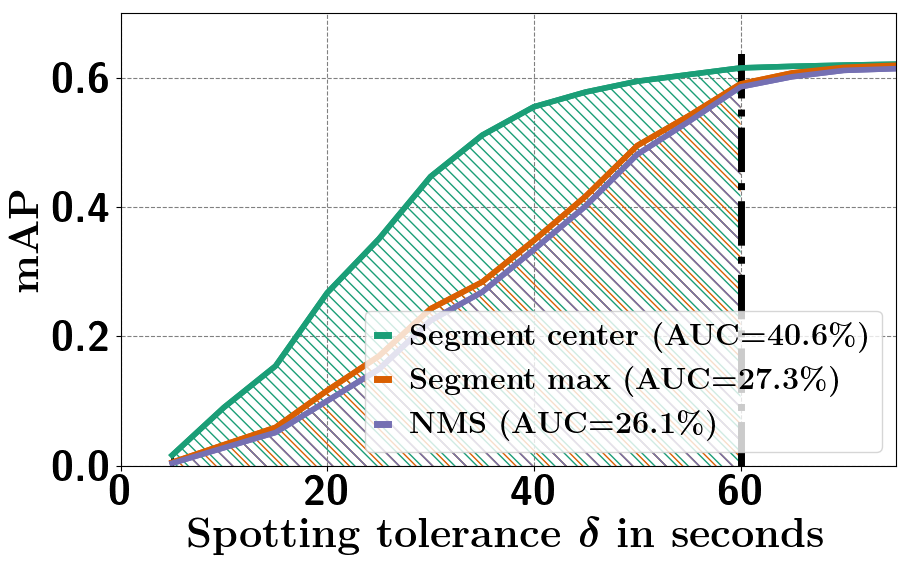}
    \label{fig:GraphSpotting_ModelVLAD512_Wind60}}
    \subfloat[\textbf{Model trained on chunks of 20s}]
    {\includegraphics[width=0.32\linewidth]{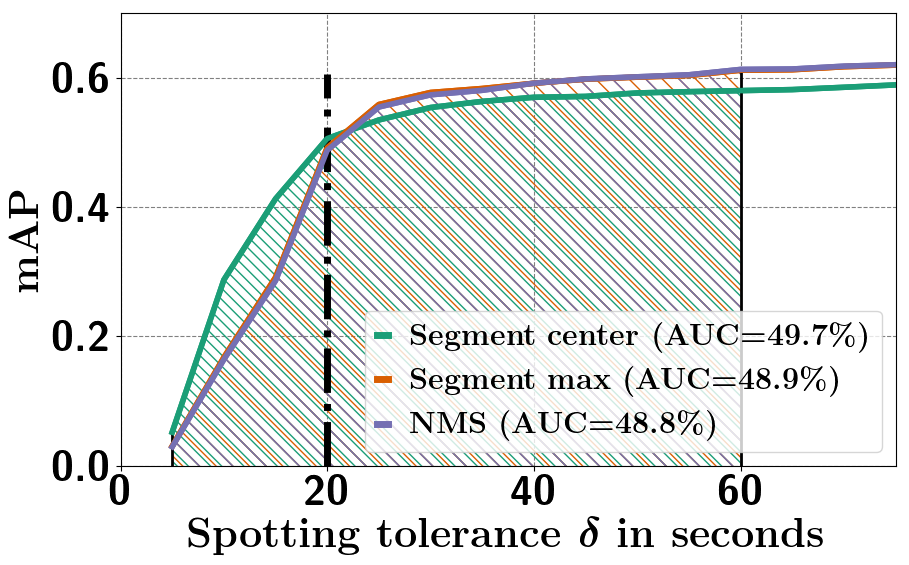}
    \label{fig:GraphSpotting_ModelVLAD64_Wind20}}
    \subfloat[\textbf{Model trained on chunks of 5s}]
    {\includegraphics[width=0.32\linewidth]{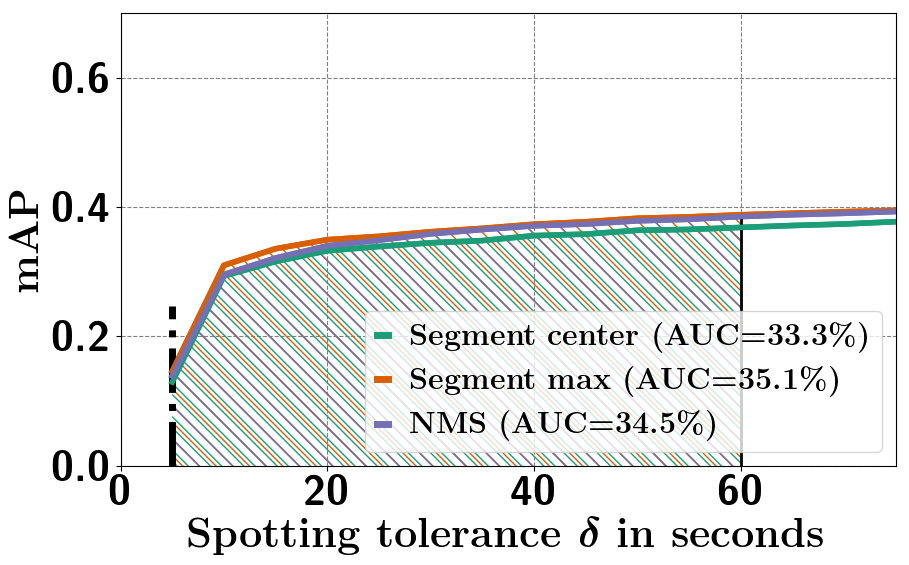}
    \label{fig:GraphSpotting_ModelVLAD64_Wind5}}
    \caption{Spotting metric (mAP) in function of the tolerance $\delta$ for model trained on chunks of size (a) 60s, (b) 20s and (c) 5s.
    The Average-mAP is estimated through the area under the curve between 5s and 60s for each baseline.
    }
    \label{fig:GraphSpotting}
\end{figure*}

\vspace{4pt}\noindent\textbf{Comparison of Spotting Baselines:~~}
We investigate the results for event spotting for our best weakly-trained classifier, to leverage the use of webly-parsed annotations, \ie we train on the imbalanced minute resolution annotated data and do not perform data augmentation.
Specifically, we use a NetVLAD model with $k=512$ cluters based on ResNet features and the watershed threshold is set to 50\%.

Figure~\ref{fig:GraphSpotting_ModelVLAD512_Wind60} plots the mAP of each spotting baseline as a function of the tolerance $\delta$ to the spot. 
As expected, the mAP decreases with the spot tolerance $\delta$. 
Above a tolerance $\delta$ of 60s, both three baselines plateau at 62.3\%.
Below 60s, the baseline (ii) and (iii) perform similarly and decrease linearly with the tolerance.
On the other hand, baseline (i) decreases more gradually, hence provides a better Average-mAP of 40.6\%.
Even though the model has been trained using chunks of 1 minute, the method is still able to achieve good spotting results for tolerances below 60s.
We argue that our model predicts positively any window that contains an event, creating a plateau.


\vspace{4pt}\noindent\textbf{Training on Smaller Windows:~~}
Here, we train our classifiers from Section~\ref{subsec:Classifier} using a smaller chunk size, ranging from 60 seconds to 5 seconds.
We expect these models to perform in a similar fashion, with a drop in performance (mAP) occurring for tolerances below the chunk size.
Note that we use finely annotated data to train such classifiers.

Figure~\ref{fig:GraphSpotting} depicts the spotting mAP in function of the tolerance $\delta$ for the models trained on 60, 20 and 5 seconds.
They all have similar shape, a metric that plateaus for spotting tolerance $\delta$ above the chunk video length they have being trained on, and a decreasing metric below such threshold.
By using baseline (i) on chunks of 20s we obtain the best Average-mAP of 50\% (see Figure~\ref{fig:GraphSpotting_ModelVLAD64_Wind20}).
Also, a drop in performance occurs with models trained with chunks of 5s (see Figure~\ref{fig:GraphSpotting_ModelVLAD64_Wind5}).
We believe such gap in performance is related to the amount of context we allow around the event.

With these experiments, we setup a baseline for the spotting task but the best performance is far from satisfactory.
Nevertheless, we see our newly compiled and scalable dataset to be a rich environment for further algorithm development and standardized evaluations; especially when it comes to novel spotting techniques.

\section{Future Work}

Activity detection is commonly solved by proposing candidates that are further classified. 
We believe that detection can be solved by spotting a candidate and focusing attention around the spot to localize the activity boundaries.

In future works, we encourage the usage of RNNs to embed a further temporal aspect that will understand the evolution of the game.
We will also include more classes for soccer events to enrich its contents and enable learning potential causal relationships between events.
We believe for instance that the event \emph{``card''} is mostly the result of an event \emph{``foul''}.
Also, embedding semantic relationship information from the players, the ball and the field can improve soccer video understanding.
Our video also contains an audio track that should be used; visual and audio sentiment analysis could localize the salient moments of the game.

The match reports from our online provider also includes match commentaries.
We collected and will release a total of 506,137 commentaries for the six aforementioned leagues with a one second resolution. 
We believe such data can be used for captioning events in soccer videos.

\section{Conclusion}

In this paper, we focus on soccer understanding in TV broadcast videos.
We build this work as an attempt to provide a benchmark for soccer analysis, by providing a large-scale annotated dataset of soccer game broadcasts.
We discussed the concept of \emph{event} within the soccer context, proposed a definition of \emph{``goal''}, \emph{``card''} and \emph{``substitution''} and parse a large amount of annotation from the web.
We defined the task of \emph{spotting} and provide a baseline for it.
For the minute classification task, we have shown performance of 67.8\% (mAP) using ResNet-152 features and NetVLAD pooling along a 512-long vocabulary and using a coarse annotation.
Regarding the spotting task, we have establish an Average-mAP of 49.7\% with fine annotation and 40.6\% by using only weakly annotated data.
We believe that focusing effort on spotting, new algorithms can improve the state-of-the-art in detection tasks.

\ifcvprfinal
\textbf{Acknowledgments.} This work was supported by the King Abdullah University of Science and Technology (KAUST) Office of Sponsored Research.
\fi

\newpage
{\footnotesize
\bibliographystyle{ieee}
\bibliography{bib/dataset,bib/egbib,bib/pooling,bib/VideoUnderstanding,bib/MarketAnalysis,bib/FootballAnalysis}
}

\cleardoublepage

\section{Supplementary Material}

We provide further details on the the dataset and further results for the spotting baseline.

\subsection{Dataset Details}

Table~\ref{tab:DatasetSplit} provides more details on the distribution of the events for the training (300 games), validation (100 games) and testing (100 games) sets. We assess that the events are equally distributed along the different sets.

\begin{table}[htb]
	\centering
	\caption{Details on the events split between Training, Validation and Testing sets.}
	\vspace{5pt}
	\begin{filecontents*}{img/Dataset/CountEventSplit.csv}
Split,goal,card,subs,Total
Train,961,1296,1708,3965
Valid,356,396,562,1314
Test,326,453,579,1358
Total,1643,2145,2849,6637
\end{filecontents*}
\csvreader[tabular=l||c|c|c||>{\bfseries}c, 
	table head=   &  \multicolumn{3}{c||}{Events} &   \\
	Split &  Goals &  Cards &  Subs &  Total \\\midrule,
	late after line=\ifthenelse{\equal{\Split}{Total}}{\\\midrule}{\\}]
	{img/Dataset/CountEventSplit.csv}%
	{Split=\Split,
	goal=\goal,
	card=\card,
	subs=\subs,
	Total=\Total}%
	{\ifthenelse{\equal{\Split}{Total}}
		{\textbf{\Split}}{\Split} & 
		\ifthenelse{\equal{\Split}{Total}}
		{\textbf{\goal}}{\goal} & 
		\ifthenelse{\equal{\Split}{Total}}
		{\textbf{\card}}{\card} & 
		\ifthenelse{\equal{\Split}{Total}}
		{\textbf{\subs}}{\subs} & 
		\ifthenelse{\equal{\Split}{Total}}
		{\textbf{\Total}}{\Total} }
	\label{tab:DatasetSplit}
\end{table}

\subsection{PCA reduction}

\begin{figure}[htb]
    \centering
    \includegraphics[width=\linewidth]{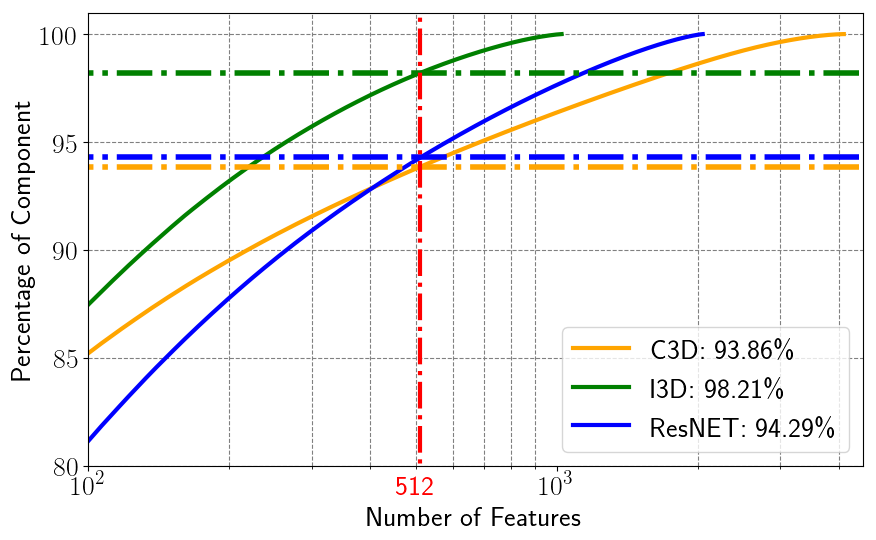}
    \caption{Dimensionality reduction using Principal Component Analysis (PCA).}
    \label{fig:supplPCA}
\end{figure}

Reducing the dimension of the frame features reduces the complexity of the successive pooling layers. Nevertheless, the features lose some variance. We ensure in Figure~\ref{fig:supplPCA} that the loss in variance is minimal when reducing the dimension to 512 for ResNET, C3D and I3D.

\subsection{Insight for the metrics}

\begin{figure}[t]
    \centering
    \includegraphics[width=\linewidth]{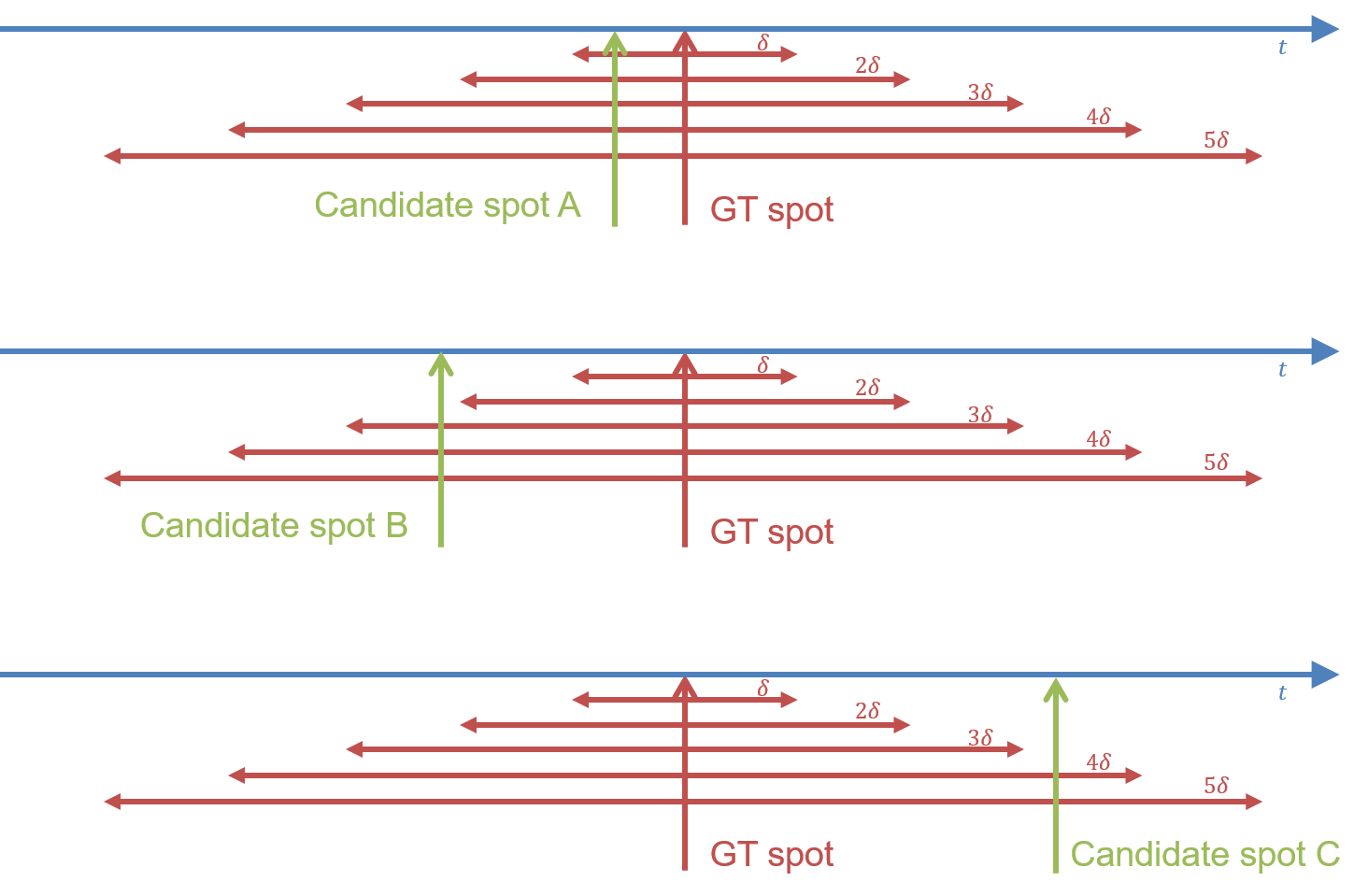}
    \caption{Insight to understand the metrics. 
    Candidate A spots the event within a tolerance of $\delta$,
    Candidate B within $3\delta$ and 
    Candidate C within $4\delta$.}
    \label{fig:supplInsightMetrics}
\end{figure}

We show in Figure~\ref{fig:supplInsightMetrics} some insight from the metric we are defining. A candidate spot is considered positive if he lands within a tolerance $\delta$ of a ground truth spot. In Figure~\ref{fig:supplInsightMetrics}, candidate A lands within a tolerance $\delta$, candidate B within a tolerance $3\delta$ and candidate C within a tolerance $4\delta$, hence considered as positive for tolerances greater or equal to such value, and negative for smaller tolerances. Recall and Precision are defined for a given tolerance $\delta$, as well as the Average Precision (AP), the mean Average Precision (mAP) along the three classes and the Average mAP along a set of tolerances. Note that the Average mAP can also be estimated through the area under the curve of the mAP in functions of the spotting tolerance (see Figure~\ref{fig:SupplGraphSpottingWindows}).

\subsection{Spotting Results}

\begin{figure}[H]
    \centering
    \includegraphics[width=\linewidth]{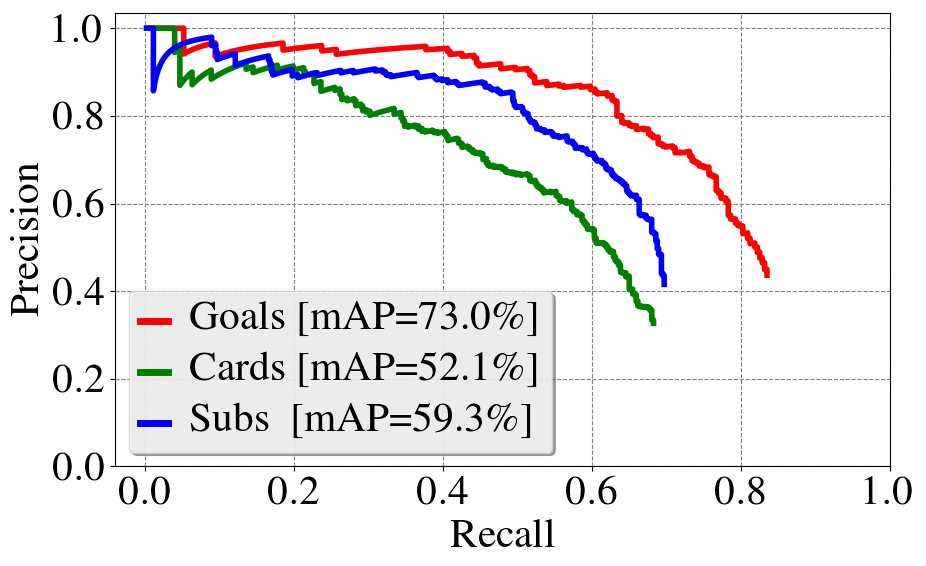}
    \caption{Recall Precision curve for the three classes.}
    \label{fig:supplRecallPrecision}
\end{figure}

We show in Figure~\ref{fig:supplRecallPrecision} the Recall Precision curve for the 3 metrics, using the best result for classification training, \ie ResNet-152, NetVLAD pooling with $k=64$ and segment center spotting baseline. Goal events are the easiest to spot with an AP of 73.0\%, then substitutions reach 59.3\% and Cards 52.1\%.

\begin{figure*}[t]
    \centering
    \subfloat[\textbf{Model trained on chunks of 50s}]
    {\includegraphics[width=0.45\linewidth]{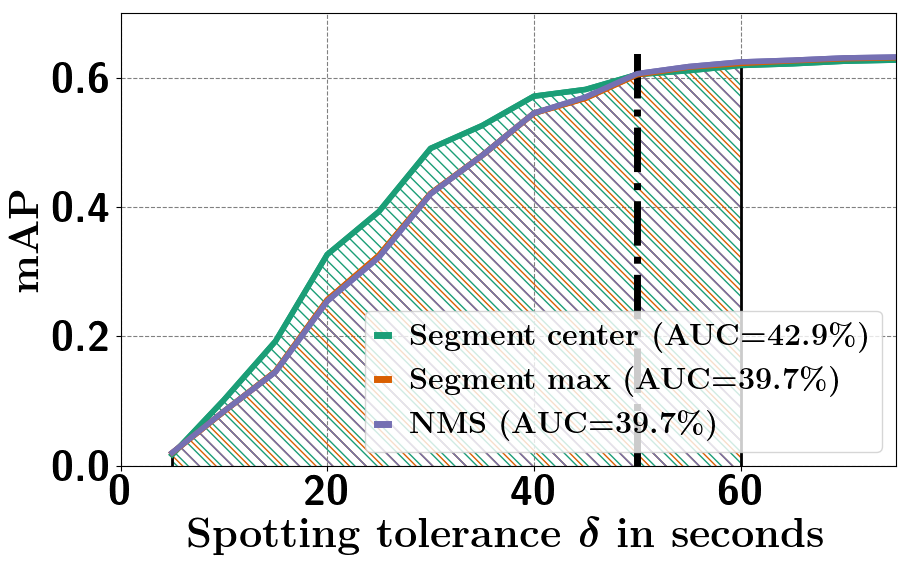}}
    \subfloat[\textbf{Model trained on chunks of 40s}]
    {\includegraphics[width=0.45\linewidth]{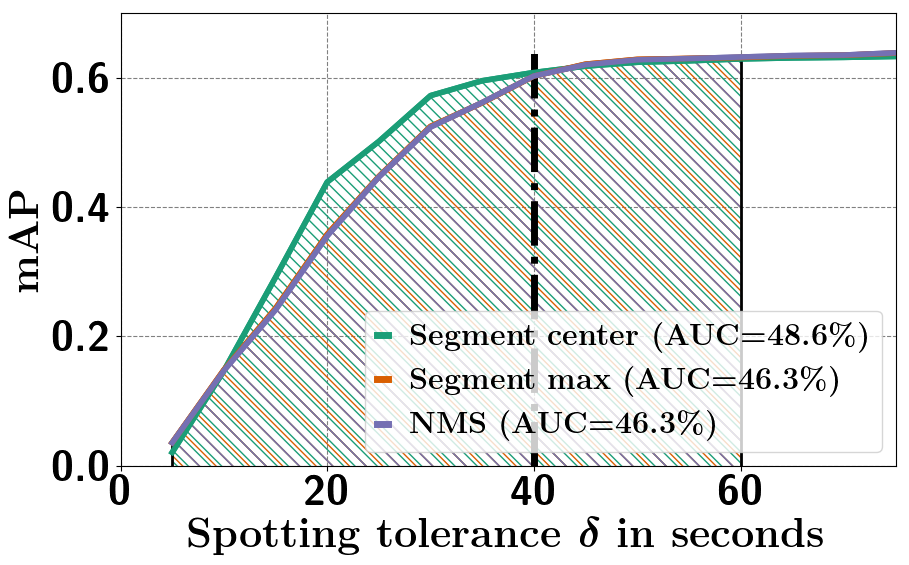}}\\
    \subfloat[\textbf{Model trained on chunks of 30s}]
    {\includegraphics[width=0.45\linewidth]{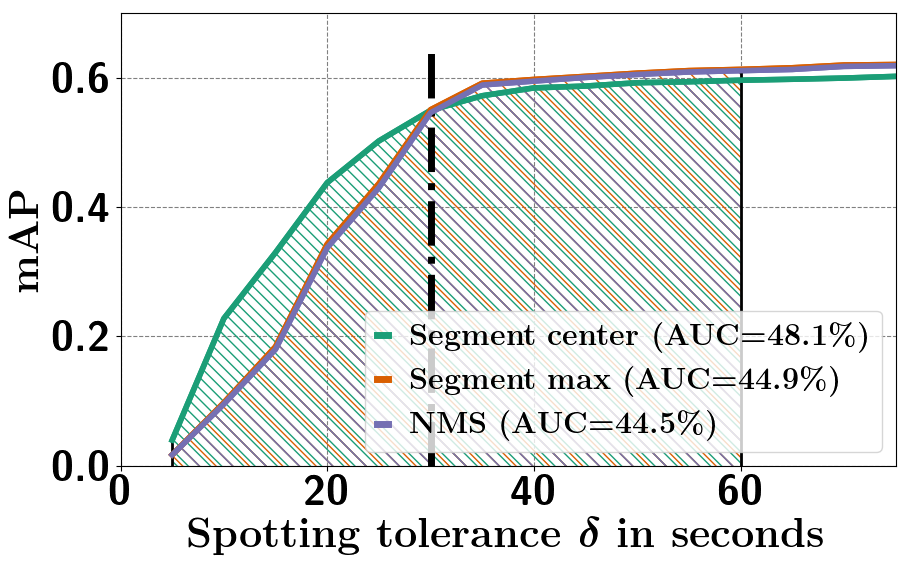}}
    \subfloat[\textbf{Model trained on chunks of 20s}]
    {\includegraphics[width=0.45\linewidth]{img/Results/GraphSpotting_ModelVLAD64_Wind20}}\\
    \subfloat[\textbf{Model trained on chunks of 10s}]
    {\includegraphics[width=0.45\linewidth]{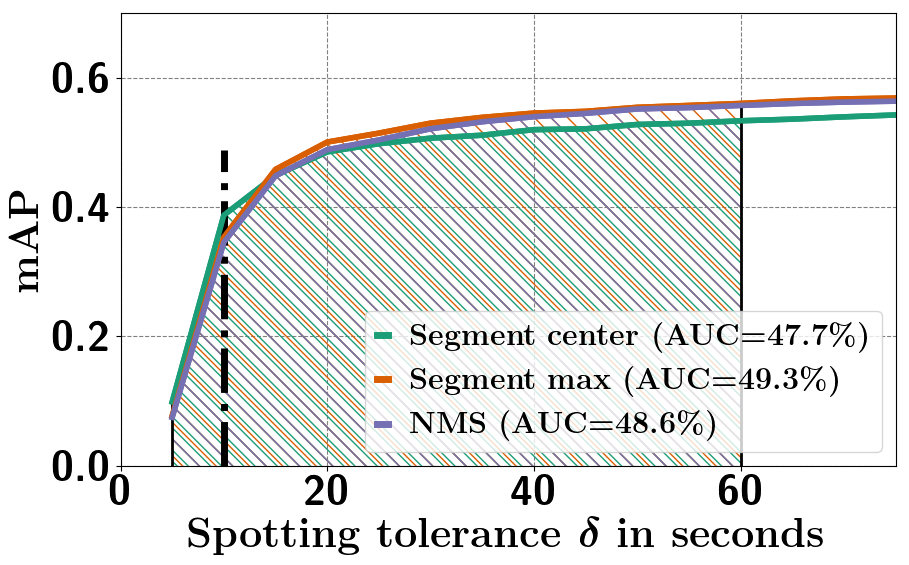}}
    \subfloat[\textbf{Model trained on chunks of 5s}]
    {\includegraphics[width=0.45\linewidth]{img/Results/GraphSpotting_ModelVLAD64_Wind5}}
    \caption{Spotting metric (mAP) in function of the tolerance $\delta$ for model trained on chunks of size (a) 50s, (b) 40s, (c) 30s, (d) 20s, (e) 10s and (f) 5s.
    The Average-mAP is estimated through the area under the curve between 5s and 60s for each baseline.}
    \label{fig:SupplGraphSpottingWindows}
\end{figure*}

Also, Figure~\ref{fig:SupplGraphSpottingWindows} illustrates the mAP and the Average mAP for different models using different window sizes during training. It shows that the best result is performed by a window size of 20 seconds in classification training. A drop in performances in visible for windows of 5 seconds. Note that such model uses the fine annotation at a one second resolution.

\subsection{Qualitative results}

\begin{figure*}[htb]
    \centering
    \subfloat
    {\includegraphics[width=\linewidth]{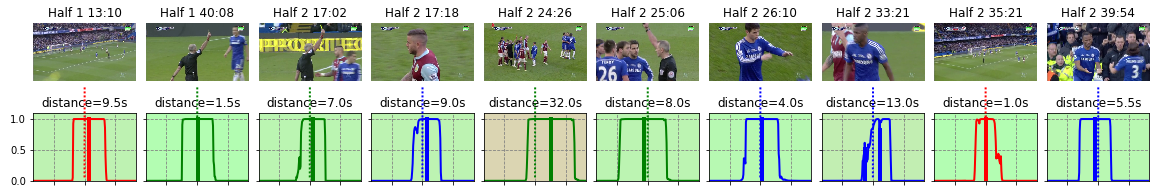}
    \label{fig:QualitativeResults_Train}}\\
    \addtocounter{subfigure}{-1}
    \subfloat[\textbf{Example of games from the training set}]
    {\includegraphics[width=\linewidth]{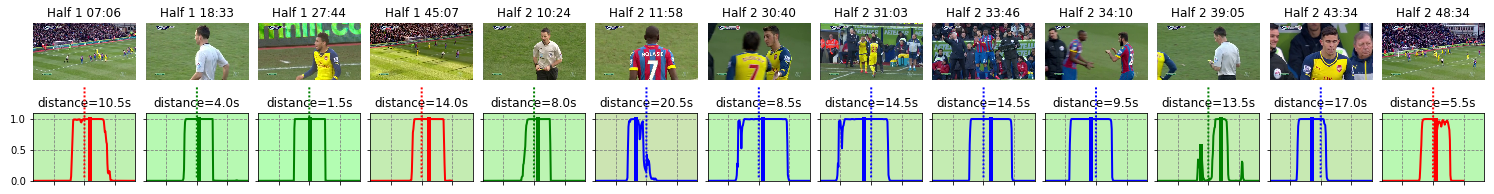}
    \label{fig:QualitativeResults_Train}}\\
    \vspace{10pt}
    \subfloat
    {\includegraphics[width=\linewidth]{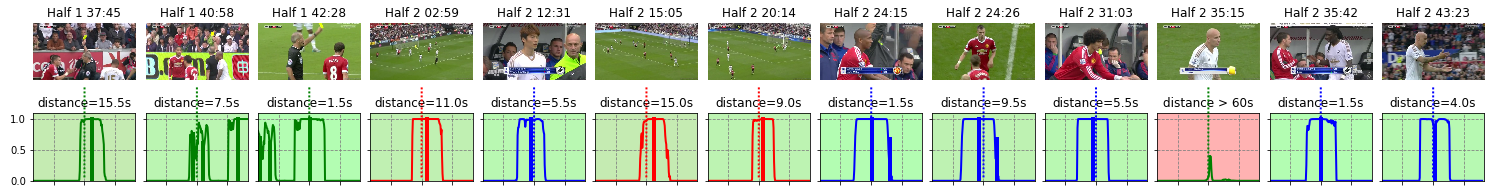}
    \label{fig:QualitativeResults_Valid}}\\
    \addtocounter{subfigure}{-1}
    \subfloat[\textbf{Example of games from the validation set}]
    {\includegraphics[width=\linewidth]{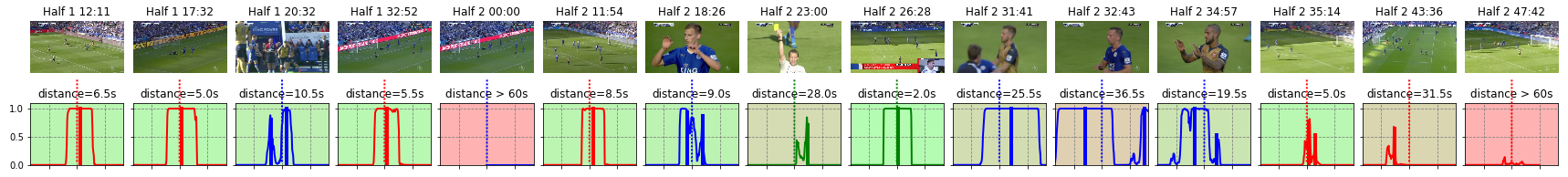}
    \label{fig:QualitativeResults_Valid}}\\
    \vspace{10pt}
    \subfloat
    {\includegraphics[width=\linewidth]{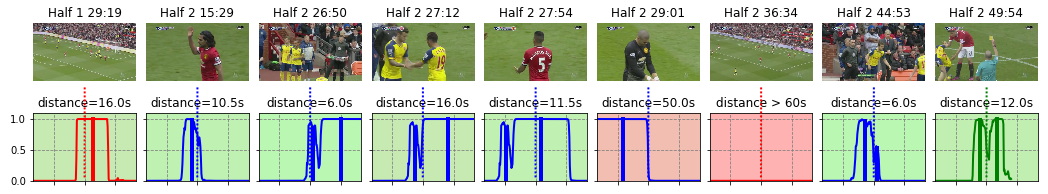}
    \label{fig:QualitativeResults_Test}}\\
    \addtocounter{subfigure}{-1}
    \subfloat[\textbf{Example of games from the testing set}]
    {\includegraphics[width=\linewidth]{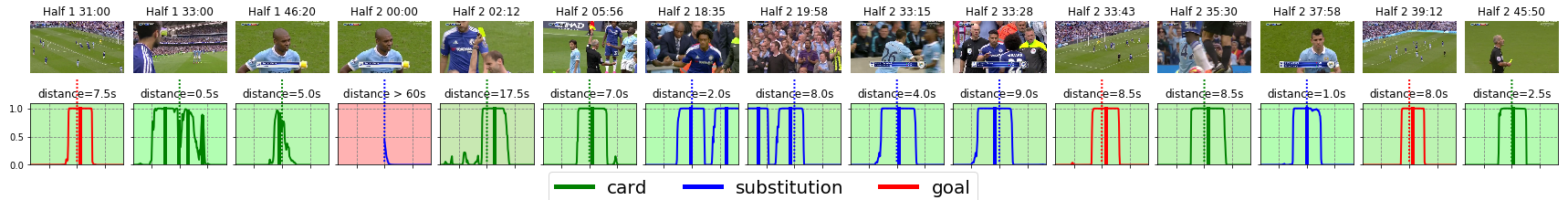}
    \label{fig:QualitativeResults_Test}}
    \caption{Qualitative results for the Training (a), Validation (b) and Testing (c) examples. The time scale (X axis) is in minute.}
    \label{fig:supplQualitativeResults}
\end{figure*}

Figure~\ref{fig:supplQualitativeResults} shows qualitative results for games in the training, validation and testing set. Each tile shows the activation around ground truth events for a given class, and depicts the candidate spot using our best model (ResNet-152, NetVLAD with $k=512$) and the center segment (i) spotting baseline.
The prediction usually activates for a 60 seconds range around the spot. It validates our hypothesis that any sliding window that contains the ground truth spot activates the prediction for the class.

\begin{figure*}[htb]
    \centering
    \includegraphics[width=\linewidth]{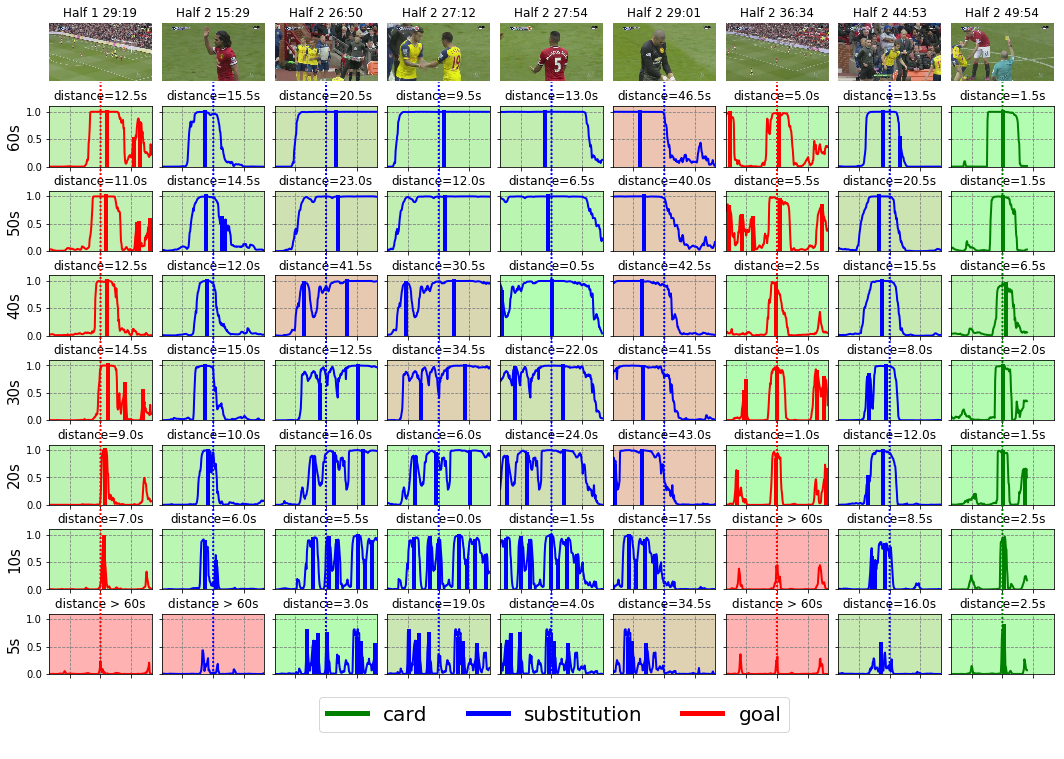}
    \caption{Qualitative results for window size ranging from 60 to 5 s.}
    \label{fig:supplQualitativeResultsWindow}
\end{figure*}

Figure~\ref{fig:supplQualitativeResultsWindow} shows further results with smaller windows sizes in training. As expected, the activation width reduces from 60 seconds to the value of the size of the video chunks used in training.

\end{document}